# SELF-ORGANISING MAPS
# IN COMPUTER SECURITY

*Jan Feyereisl[1] and Uwe Aickelin*
The University of Nottingham
Nottingham, UK

### Abstract

Some argue that biologically inspired algorithms are the future of solving difficult problems in computer science. Others strongly believe that the future lies in the exploration of mathematical foundations of problems at hand. The field of computer security tends to accept the latter view as a more appropriate approach due to its more workable validation and verification possibilities. The lack of rigorous scientific practices prevalent in biologically inspired security research does not aid in presenting bio-inspired security approaches as a viable way of dealing with complex security problems. This chapter introduces a biologically inspired algorithm, called the SelfOrganising Map (SOM), that was developed by Teuvo Kohonen in 1981. Since the algorithm's inception it has been scrutinised by the scientific community and analysed in more than 4000 research papers, many of which dealt with various computer security issues, from anomaly detection, analysis of executables all the way to wireless network monitoring. In this chapter a review of security related SOM research undertaken in the past is presented and analysed. The algorithm's biological analogies are detailed and the author's view on the future possibilities of this successful bio-inspired approach are given. The SOM algorithm's close relation to a number of vital functions of the human brain and the emergence of multi-core computer architectures are the two main reasons behind our assumption that the future of the SOM algorithm and its variations is promising, notably in the field of computer security.

## 1. Introduction

"Nothing in security really works!" A recurring theme during a panel discussion on biologically inspired security that summarises current state of the security field [99]. The security community frequently argues that approaches stemming from the biological realm are a frequent source of poor science or research that is not applicable to the real world. Nevertheless the fact that the community itself has trouble finding answers to many prevailing problems is testament to the need of the security field to look beyond traditional means of solving problems.

---

[1] E-mail address: jqf@cs.nott.ac.uk

The issue of security has been pursuing every species on our planet since life began. The survival of any species is based on its ability to ensure its own security. Over the millennia different species have evolved and learned numerous techniques to increase the level of security that pertained to their survival. Man evolved gestures, better physical stamina, invented fences, weapons, law and many other tools and techniques that enabled him to keep up with the world around him. In the last fifty years however, the explosive nature of the digital age opened up new challenges that have never been dealt with before. The creation of complex systems that have been develop by us, in many cases for purposes other than security, are now increasingly being misused exactly for that purpose. To exploit the insecure nature of these devices and their possible gain to the malicious users.

The digital security field, as we know it today, has started with the creation of cryptographic protocols that have been used to transfer military secrets during the second world war. Since then computers have become increasingly part of everyday life and security focus has shifted from specialised applications to more mainstream, business oriented protection of assets and data. In the last decade this focus has also broadened into the area of personal computing where the lack of knowledge of digital systems by their users provides easy target for attackers.

Numerous different techniques have been devised over the years for the purpose of detecting and stopping intruders, identifying malicious users, categorising malicious behaviour and dealing with all types of illegal or rogue activities in the digital realm. These range from user-centric approaches, such as educating the users about possible threats that can be encountered within the digital world, to techno-centric ones where mathematical, engineering and other technological methods are employed to tackle the various security issues.

In this chapter we will focus on the introduction of an approach that has stemmed from a biological inspiration, yet is based on strong mathematical foundations, that gave it a number of properties suitable for various security purposes. This algorithm, developed by Teuvo Kohonen in 1981, is called the Self-Organising Map (SOM) [55]. It has found a wide audience across many disciplines of computer science, including security. We will describe its functionality, its advantages as well as disadvantages, the algorithm's variations and present work that has been undertaken in order to exploit the algorithm's capabilities in the computer security field. A discussion of the algorithm's possible future, with references to state of the art hardware as the underlying mechanism to push the algorithm's capabilities in real-world applications, will conclude the chapter.

## 2.   The SOM Algorithm

The Self-Organising Map algorithm was developed more than two decades ago [55], yet its success in various fields of science, over the years, surpasses many other neural inspired algorithms to date. The algorithm's strengths lie in a number of important scientific domains. Namely visualisation, clustering, data processing, reduction and classification. In more specific terms SOM is an unsupervised learning algorithm that is based on the competitive learning mechanism with self-organising properties. Besides its clustering properties, SOM can also be classed as a method for multidimensional scaling and projection.

## 2.1.   SOM as a Biological Inspiration

Various properties of the brain were used as an inspiration for a large set of algorithms and computational theories known as neural networks [38]. Such algorithms have shown to be successful, however a vital aspect of biological neural networks was omitted in the algorithm's development. This was the notion of self-organisation and spatial organisation of information within the brain. In 1981 Kohonen proposed a method which takes into account these two biological properties and presented them in his SOM algorithm [55].

The SOM algorithm generates, usually, two dimensional maps representing a scaled version of $n$-dimensional data used as the input to the algorithm. These maps can be thought of as "neural networks" in the same sense as SOM's traditional rivals, artificial neural networks (ANNs). This is due to the algorithm's inspiration from the way that mammalian brains are structured and operate in a data reducing and self-organised fashion. Traditional ANNs originated from the functionality and interoperability of neurons within the brain. The SOM algorithm on the other hand was inspired by the existence of many kinds of "maps" within the brain that represent spatially organised responses. An example from the biological domain is the somatotopic map within the human brain, containing a representation of the body and its adjacent and topographically almost identical motor map responsible for the mediation of muscle activity [58].

This spatial arrangement is vital for the correct functioning of the central nervous system [47]. This is because similar types of information (usually sensory information) are held in close spatial proximity to each other in order for successful information fusion to take place as well as to minimise the distance when neurons with similar tasks communicate. For example sensory information of the leg lies next to sensory information of the sole.

The fact that similarities in the input signals are converted into spatial relationships among the responding neurons provides the brain with an abstraction ability that suppresses trivial detail and only maps most important properties and features along the dimensions of the brain's map [91].

## 2.2.   Algorithmic Detail

As the SOM algorithm represents the above described functionality, it contains numerous methods that achieve properties similar to the biological system. The algorithm comprises of competitive learning, self-organisation, multidimensional scaling, global and local ordering of the generated map and its adaptation.

There are two high-level stages of the algorithm that ensure a successful creation of a map. The first stage is the *global ordering* stage in which we start with a map of predefined size with neurons of random nature and using competitive learning and a method of self-organisation, the algorithm produces a rough estimation of the topography of the map based on the input data. Once a desired number of input data is used for such estimation, the algorithm proceeds to the *fine-tuning* stage, where the effect of the input data on the topography of the map is monotonically decreasing with time, while individual neurons and their close topological neighbours are sensitised and thus fine tuned to the present input.

The original algorithm developed by Kohonen comprises of initialisation followed by three vital steps which are repeated until a condition is met:

- *Choice of stimulus*

- *Response*

- *Adaptation*

Each of these steps are described in detail in the following sections.

### 2.2.1.   Initialisation

A number of parameters have to be chosen before the algorithm is to begin execution. These include the size of the map, its shape, the distance measure used for comparing how similar nodes are, to each other and to the input feature vectors, as well as the kernel function used for the training of the map. Kohonen suggested recommended values for these parameters [58], nevertheless suitable parameters can also be obtained experimentally in order to tailor the algorithm's functionality to a given problem. Once these parameters are chosen, a map is created of the predefined size, populated with nodes, each of which is assigned a vector of random values, $w_i$, where $i$ denotes node to which vector $w$ belongs.

### 2.2.2.   Stimulus Selection

The next step in the SOM algorithm is the selection of the stimulus that is to be used for the generation of the map. This is done by randomly selecting a subset of input feature vectors from a training data set and presenting each input feature vector, $x$, to the map, one item per epoch. An epoch represents one complete computation of the three vital steps of the algorithm.

### 2.2.3.   Response

At this stage the algorithm takes the presented input $x$ and compares it against every node $i$ within the map by means of a distance measure between $x$ and each nodes' weight vector $w_i$. For example this can be the Euclidean distance measure shown in Equation 1, where $||.||$ is the Euclidean norm and $w_i$ is the weight vector of node $i$. This way a winning node can be determined by finding a node within the map with the smallest Euclidean distance from the presented vector $x$, here signified by $c$.

$$c = argmin\{||x - w_i||\} \tag{1}$$

### 2.2.4.  Adaptation

Adaptation is the step where the winning node is adjusted to be slightly more similar to the input $x$. This is achieved by using a kernel function, such as the Gaussian function ($h_{ci}$) as seen in Equation 2 as part of a learning process.

$$h_{ci}(t) = \alpha(t).exp\left(-\frac{||r_c - r_i||^2}{2\sigma^2(t)}\right) \tag{2}$$

In the above function $\alpha(t)$ denotes a "learning-rate factor" and $\sigma(t)$ denotes the width of the neighbourhood affected by the Gaussian function. Both of these parameters decrease monotonically over time ($t$). During the first 1,000 steps, $\alpha(t)$ should have reasonably high values (e.g. close to 1). This is called the *global ordering* stage and is responsible for proper ordering of $w_i$. For the remaining steps, $\alpha(t)$ should attain reasonably small values ($\geq 0.2$), as this is the *fine-tuning* stage where only fine adjustments to the map are performed. Both $r_c$ and $r_i$ are location vectors of the winner node (denoted by subscript $c$) and $i$ respectively, containing information about a node's location within the map.

$$w_i(t + 1) = w_i(t) + h_{ci}(t)[x(t) - w_i(t)] \tag{3}$$

The learning function itself is shown in Equation 3. Here the Gaussian kernel function $h_{ci}$ is responsible for the adjustment of all nodes according to the input feature vector $x$ and each node's distance from the winning node. This whole adaptation step is the vital part of the SOM algorithm that is responsible for the algorithm's self-organisational properties.

### 2.2.5.  Repetition

Stimulus selection, Response and Adaptation are repeated a desired number of times or until a map of sufficient quality is generated. Kohonen [57] states that the number of steps should be at least 500 times the number of map units. Another possible mechanism for the termination of the algorithm is the calculation of the quantisation error. This is the mean of $||x - w_c||$ over the training data. Once the overall quantisation error falls below a certain threshold, the execution of the algorithm can stop as an acceptable lower dimensional representation of the input data has been generated.

## 2.3.    Variations of the SOM Algorithm

Kohonen's original incremental SOM algorithm was the first in a series of algorithms based on the idea of maps created by the process of self-organisation for the purpose of visualisation, clustering and dimensionality reduction. Kohonen proposed a number of improvements to his original algorithm, such as the "Batch SOM" [58] as well as the "Dot-Product SOM" [58] and most recently a SOM which identifies a linear mixture of model vectors instead of winner nodes [59]. Kusumoto [63] proposed a more efficient SOM algorithm called $0(log_2M)$, which introduced a new method of self-organisation based on a sub-division technique which inherently deals with information propagation to neighbourhood nodes within the generated map. Due to the unique structured approach of self-organisation, the search for the "winner" neurons can be performed using a binary search, which greatly enhances the performance of the algorithm [64]. Berglund and Sitte [6] proposed a *Parameterless* SOM, where the problem of selection of suitable learning rate and annealing scheme is solved, however at the cost of introduction of some errors with the topology preservation of the generated map.

The above mentioned approaches are mainly improvements in terms of optimisation and data representation. Other algorithms which attempt to extend or alter Kohonen's original idea in a more significant manner were proposed by a number of other researchers. These include the *Hierarchical* SOM [49], which contains an additional layer of maps, linked to and generated from nodes within the original map, where the node's activation level exceeds a predefined threshold or other approaches that attempt automatic determination of the map's size, based on the properties of the original algorithm. These include the *Growing Grid* SOM [28] and *Growing Neural Gas* [27] algorithms. Such algorithms start with a minimal size of the map (usually 2x2 nodes), which over the duration of execution incrementally grows as and when required by the input data. These algorithms present some advantages in comparison with the original SOM, for example improved data representation as well as memory and speed optimisations, however they also bring some drawbacks, such as issues with visualisation.

## 2.4.    SOM in Computational Intelligence

From a computational intelligence point of view, the strengths of the SOM algorithm lie particularly in three areas. First of all due to the fact that the SOM algorithm generates a lower dimensional feature map, the algorithm is suitable for *visualising* multi-dimensional and complex data in a way that enables better understanding of such data. Secondly the self-organisational properties and topological preservation provide a way for data to be organised in *clusters*. This also aids in visualising the relationship between the observed data as well as the possibility to use this knowledge for many computational intelligence problems, such as anomaly/novelty detection or general exploratory data analysis within data mining. Ultsch and Siemon devised a technique, called the *Unified Distance Matrix* (U-Matrix), to meaningfully represent a feature map generated by a SOM algorithm [100]. This technique is now the de facto standard for visualising SOM feature maps. The SOM algorithm on its own is first and foremost regarded as a visualisation and clustering algorithm, nevertheless with additional steps added at the post-processing stage, the algorithm can also be used as a *classification* tool. Kohonen however suggests to use the *Learning Vector Quantisation* (LVQ) algorithm, which is more suited for this task [56].

The use of the SOM algorithm generally falls into one of the three above mentioned categories. In the next section we will refer to this categorisation in order to distinguish the use of the algorithm within the security field.

## 3. Self-Organising Map and Security

The SOM algorithm has been applied to many different areas of computer security in the past. There are over a hundred research papers written on this topic, where the SOM algorithm is used to solve or aid another technique in dealing with a security problem. In the rest of this chapter we will describe existing research, evaluate the algorithm's impact on the field and provide pointers for future research within this area.

This section is structure based on existing security problem areas. We start with the description of the most researched area, software security, followed by the application of the algorithm in the more tactile area of security, hardware security. Other security problems also tackled using SOM, such as forensics and cryptography follow. The section ends with the description of application of the SOM algorithm within the more exotic, or difficult to classify, areas of security, such as home security.

### 3.1. Software Security

SOM algorithms have been first applied to computer security applications almost ten years after the algorithm's inception [26]. The majority of existing research however is limited to anomaly detection, particularly network based intrusion detection. Some work has been done on host based anomaly detection using Kohonen's algorithm, however such work is still rare, which is surprising, due to the algorithm's suitability to handle multidimensional, thus multi-signal data. On numerous occasions SOM algorithms have been used as a pre-processor to other computational intelligence tools, such as Hidden Markov Models (HMM) [15] [14] [52] or Radial Basis Function (RBF) Networks [42]. Comparisons of SOM algorithm with other anomaly detection approaches have been performed on numerous occasions in the past. Notably a comparison with HMM [104], Artificial Immune Systems (AIS) [32] [33], traditional neural networks [93] [66] [46] [65] [8] as well as Adaptive Resonance Theory (ART) [2].

Besides anomaly and intrusion detection, the SOM algorithm has been applied to binary code analysis for the purpose of virus, payload and buffer overflow detection as well as attack and vulnerability characterisation and classification. Alert filtering and correlation are also areas that benefit from the capabilities of the SOM algorithm. There are many other software security areas that Kohonen's algorithm has been applied to. These are described in detail in the following section.

### 3.1.1.   Intrusion and Anomaly Detection

The field of intrusion and anomaly detection (IDS) has been one of the most actively researched areas of security for many years. There are a number of different types of intrusion detection systems, depending on their functionality and approach with which they deal with intrusions and anomalies. There are two high level categorisations of such systems. The first category group being signature and anomaly based systems. These two categories of systems differ in the way that they hold knowledge about possible intrusions. *Signature* based systems contain a database of generated signatures which are used to recognise existing malicious entities. *Anomaly* based systems on the other hand hold a baseline of normal behaviour of a system, which is used to recognise if a system's behaviour somehow deviates from this baseline. For a more detailed definition of such systems, please refer to [31]. A general overview of novelty detection using neural networks including SOM can be found in [73].

Anomaly Based Systems The majority of systems described in the following sections are anomaly based. This is mainly due to the fact that the SOM algorithm enables the creation of a baseline suitable for such types of systems.

Signature Based Systems There are only a few systems that can be thought of as signature based in the traditional sense. All of these systems are hybrid systems, which combine both anomaly as well as signature based techniques in order to achieve the best possible detection capabilities. An example of such a system was developed by Powers and He [85]. In their work the SOM algorithm is used to generate higher level description of attack types which are subsequently used to classify anomalous connections detected by an anomaly detector. Another example is work by Depren *et al.* [18], who use SOM as an anomaly detector in combination with a decision tree algorithm called J.48 used as a misuse detector. In their work it is shown that the combined system has a better detection performance than the algorithms individually on their own.

The second category group distinguishes systems based on what type of information they monitor. These systems can be categorised into *network* and *host* based detection systems.

Network Based Systems As mentioned earlier, the majority of research done using the SOM algorithm has been based on network intrusion detection. In general such work is based on the observation of various features of network packets and their impact on the detection of malicious network traffic or behaviour. In this section we will provide an overview of some network based IDS systems that have used SOM.

The majority of IDS-based work has been tested on a number of seminal datasets developed by the DARPA Intrusion Detection Evaluation Program in 1998, 1999 and 2000. The 1998 dataset has been used for the challenge of the Fifth International Conference on Knowledge Discovery and Data Mining (KDD'99). The following research work related to the SOM algorithm has been tested using this dataset [68, 50, 49, 48, 92, 103, 76, 110, 43, 67, 70, 83, 46, 18, 33, 42, 93, 81]. Besides network data the 1999 dataset contains a

small set of system data, namely file system data, however this is not always used in experiments. This dataset has been used in the following work [111, 112, 9, 33, 52, 98]. The 2000 dataset has thus far not been employed in the context of SOM research. These datasets have been heavily criticised in the past [75], nevertheless they are still the only available datasets that can be used to some extent for the purpose of comparison of various security research.

Besides these datasets, a number of research work has been tested either on synthetic or real world datasets created by authors themselves [2, 8, 48]. For example Kayacik and Zincir-Heywood [48] state that their framework for creating synthetic data for security testing purposes can generate data that is more similar to real-world data than the KDD'99 dataset. They use SOM in order to compare the two datasets and determine which dataset is more suitable for real-world security testing.

Some work has also been tested on real-life scenarios as part of an existing network. An example of such system is a seminal paper on the use of SOM algorithms for intrusion detection by Ramadas *et al.* [89]. Their work employs the original SOM algorithm as a network based anomaly detection module for a larger IDS. Besides being able to monitor all types of network traffic including SMTP protocol, the authors state that the SOM algorithm is particularly suitable for the detection of buffer overflow attacks. However, as with the majority of anomaly detection systems, the algorithm struggles to recognise attacks which resemble normal behaviour in addition to boundary case behaviour, giving rise to false positives. Another example is the work of Rhodes [90], who monitors requests to Domain Name Service (DNS) ports in order to also detect buffer overflows. In this work only TCP traffic is observed.

Other interesting network based research using Kohonen's algorithm includes the work of Amini *et al.* [2], who developed a real-time system for the monitoring of TCP/UDP and ICMP packets. In their work SOMs are combined with Adaptive Resonance Theory (ART) networks, which were found to be better than SOM. Amini's work includes time as one of the input attributes, which is said to be vital for denial of service (DoS) detection.

Bivens *et al.* [8] also test their system against DoS as well as distributed denial of service (DDoS) attacks and portscans. In their work SOM is used as a clustering method for multilayer perceptrons (MLP). By using SOM, it is possible to scale down a dynamic number of inputs into a preset lower dimensional representation. Jirapummin *et al.* [46] use SOM for the detection of SYN flooding and port scanning attacks. In their work SOM is used as a first layer into an *resilient propagation* neural network (RPN).

Other researchers also attempt to detect DoS attacks. For example Mitrokotsa and Douligeris [77] use an improved version of Kohonen's SOM algorithm called *Emergent* SOM (ESOM) where the created feature map is not limited to a small number of nodes. The advantage of using ESOM is the automatic creation of higher level structures that cannot be created using the original SOM algorithm. On the other hand the fact that the size of the created feature map is usually large, means that the computational overhead is too large for real world scenarios. Li *et al.* [67] use another extended version of the SOM algorithm, however in this case to detect DDoS attacks. Their findings show that their extended SOM algorithm surpasses Kohonen's original algorithm in DDoS detection.

Host Based Systems Host based intrusion detection systems do not appear in such abundance as network based systems, nevertheless this area of intrusion detection is becoming more active in the last few years. In host based intrusion detection, attributes other than only network features are observed in order to detect intrusions. These can include system specific signals, such as file usage, memory usage and other host based indicators.

For example Wang *et al.* [104] use the University of New Mexico live FTP dataset, which contains system call information about running processes on a system, as well as their own system call based dataset from a university network. In their work they compare the SOM algorithm with a HMM method. Their conclusion is that focusing on the transition property of events, used within HMM, can yield better results than focusing on the frequency property of events, used for their SOM. Nevertheless their work uses data which does not contain many dimensions.

On the other hand the work of Wang *et al.* [102] attempts to perform host based intrusion detection using system data with many dimensions. In their work three layers of system signals are used, system layer, process layer and network layer. A feature map is generated for each layer, thus a total of 21 different host and network based signals are used as input into the SOM algorithm. Wang and colleagues conclude that their work shows promising results, nevertheless a sensitivity analysis has to be performed in order to select the most suitable parameters.

Hoglund *et al.* [40] use SOM in order to monitor user behaviour in a real-life UNIX environment. A total of 16 different host based features are chosen as input into the SOM algorithm. Their results are encouraging however they state that the system is susceptible to false positives as well as the possibility of the system to gradually adapt to attacks if deviations are not dealt with immediately.

Cho [14] uses various host based features, such as system calls, file access and process information in order to perform intrusion detection using a hybrid system, which employs SOM, HMM and fuzzy logic. In this system, SOM determines the optimal measure of audit data and performs a data reduction function in order to be able to feed the audit data into a HMM model. Cho's conclusion is that the combination of soft- and hard-computing techniques can be successfully combined for the purpose of intrusion detection.

Lichodzijewski *et al.* [69] develop a hierarchical SOM based intrusion detection system that focuses on monitoring host "session information". The authors state that this method has a significant advantage over traditional system audit trail approaches in terms of smaller computational overhead. Another important remark in this work is the finding that an implicit method for representing time, which has no knowledge of time of day, is able to provide a much clearer identification of abnormal behaviour in comparison to a method which has explicit knowledge of time. "Session activity" is also used by Khanna and Liu [52] who use other host based indicators such as system calls, CPU, network and process activity as well.

Hybrid Approaches Besides Kohonen's algorithm, many approaches to intrusion detection exist. A number of researchers attempted to extract the best features of two or more approaches to intrusion detection and combine them in order to increase their performance. For example Albarayak *et al.* [1] proposed a unique way of combining a number of existing SOM approaches together in a node based IDS. Their thesis is of automatically determining the most suitable SOM algorithm incarnation for each node within their system. Such a decision can be achieved using heuristic rules that determine the most suitable SOM algorithm based on the nodes' environment.

Miller and Inoue [76] on the other hand suggest using multiple intelligent agents, each of which contains a SOM on its own. Such agents combine a signature and anomaly based detection technique in order to achieve a collaborative IDS, which is able to improve its detection capabilities with the use of reinforcement learning.

A number of researchers combine SOM with other neural network approaches. For example Jirapummin *et al.* [46] use SOM as a first layer into a *resilient propagation* neural network. Sarasamma and Zhu [93] use a *feedforward* neural network in order to create a *hyper-ellipsoidal* SOM which generates clusters of maximum intra-cluster and minimum inter-cluster similarity in order to enhance the algorithm's classification ability. Kumar and Devaraj [61] combine SOM with a *back propagation* neural network (BPN) for the purpose of visualising and classifying intrusions. Lee and Heinbuch [65] use SOM as part of a *hierarchical* neural network approach where SOM is used as an anomaly classifier. The authors state that their approach is 100% successful in detecting specific attacks without *a priori* information about the attacks.

Horeis [42] combines SOM with RBF networks. His results show that the combination of the two approaches provides better results than RBF itself at the expense of larger computational overhead. Horeis describes human expert integration within his system, which provides for fine-grained tuning of the system based on expert knowledge. Pan and Li [83] also combine SOM with RBF in order to determine the optimal network architecture of the RBF network for the purpose of novel attack detection.

Carrascal *et al.* [12] combine the SOM algorithm with Kohonen's classification, LVQ, algorithm. In their work SOM is used for traffic modelling, while LVQ is used for final network packet classification.

Support Vector Machines have also been used in the past. Both Khan *et al.* [51] and Shon and Moon [97] use SVMs for the purpose of anomaly detection along with SOM. Khan *et al.* [51] use SVM for classification, while employing dynamically growing selforganising tree for clustering, for the purpose of finding boundary data points between classes that are most suitable for the training of SVM. This approach is said to improve the speed of the SVM training process. Shon and Khan [97] on the other hand use SOM as part of an enhanced SVM for the purpose of packet profiling and normal profile generation. Their enhanced SVM system is compared to existing signature based systems and have shown comparable results, however with the advantage that no *a priori* knowledge of attacks is given to the enhanced SVM system, unlike the signature based systems.

Hidden Markov Models have been used on numerous occasions [15, 52, 14]. Choy and Cho [15] use SOM as a data reduction tool for raw audit data which is subsequently used for normal behaviour modelling of users using HMM. In this work it has been shown that modelling of individual users surpasses modelling of groups of users in terms of performance as well as detection ability. In the work of Khanna and Liu [52] a supervised

SOM is again used as a data reduction tool for creating more suitable input for HMM. Their HMM method is used to predict an attack that exists in the form of a hidden state. Cho [14] uses a combination of SOM, HMM and fuzzy logic, where SOM acts again as a data reduction tool necessary for the functionality of HMM.

Other hybrid approaches include a combination of SOM with a *decision tree algorithm* (DTA) [18], AIS approaches such as the one developed by Powers and Hu [85] and Gonzales *et al.* [34] as well as a combination with *Bayesian belief networks* [21], *principal component analysis*(PCA) [4] or *genetic algorithms* (GA) [72].

Depren's [18] work employs a DTA called J.48 in order to create a hybrid anomaly and misuse detection system. Powers and Hu [85] developed a system with similar intentions, however in this case the authors combine the SOM algorithm with an AIS algorithm called Negative Selection. Another AIS based approach was developed by Gonzales *et al.* [34]. In their work the SOM algorithm is also combined with the Negative Selection algorithm, but rather than used only as a classification tool it is also used for the visualisation of self/non-self feature space. This visualisation enables the understanding of the space that contains normal as well as both known and unknown abnormal. Faour *et al.* [21] use a combination of SOM and Bayesian belief networks in order to automatically filter intrusion detection alarms. Bai *et al.* [4] introduce PCA as a method for feature selection, while a multi-layered SOM is used to enhance clustering of a single SOM for the purpose of anomaly detection. The authors state that PCA reduces computational complexity and in combination with SOM provides suitable functionality as a classifier for intrusion detection.
Ma [72] suggests the use of a GA to create a genetic SOM. In this model the GA is used to train the synaptic weights of the SOM. Ma's results show that this method can be used as a clustering method, however at present time only on small-scale datasets. Another issue with this system being the necessity of *a priory* knowledge of cluster count.

From the available research it is apparent that hybrid approaches generally superseed the performance of systems based on only one method. The SOM algorithm, whether used as a clustering, visualisation or classification tool, does bring advantages to other intrusion detection methods in terms of better performance, easier understanding of the problem or better detection capabilities.

Hierarchical Approaches A number of papers discuss the advantages of using multiple or hierarchical SOM networks in contrast to a single network SOM. These include the work of Sarasamma *et al.* [94], Lichodzijewski *et al.* [68, 69] and Kayacik *et al.* [49, 50] who all use various versions of the *Hierarchical* SOM or employ multiple SOM networks for the purpose of intrusion detection. Kayacik *et al.* [49] state that the best performance is achieved using a 2-layer SOM and that their results are by far the best of any unsupervised learning based IDS to date.

As mentioned earlier Albarayak *et al.* [1] propose a method for combining different SOM approaches based on their suitability for a particular problem. In their model different SOM algorithms are implemented at different layers.

Rhodes *et al.* [90] develop a system which combines three Kohonen maps, each of them for a separate protocol. The authors argue that it would be unreasonable for a single Kohonen map to usefully characterise information from all three protocols. Their results show encouragement for their method, however they state that even a single map is able to detect anomalous features of a buffer overflow attack. Their claims are however not statistically proven.

A similar approach was taken by Wang *et al.* [102]. In their work the authors also create three SOM maps, each of which represents one of the following layers, system, process and network. Their results are also said to be encouraging, nevertheless a more thorough sensitivity analysis has to be performed first in order to tune the system to an acceptable level.

Khan and colleagues [51] use a hierarchical approach based on a dynamically growing self-organising tree in order to perform clustering for the purpose of finding most suitable support vectors for an SVM algorithm.

Comparison with Other Approaches Some researchers attempted to compare and contrast SOM based approaches with other established IDS techniques. Gonzalez and Dasgupta [32] for example compare SOM against an AIS algorithm. Their Real-Valued Negative Selection algorithm is based on the original Negative Selection algorithm proposed by Forrest *et al.* [25] with the difference of using a new representation. The original Negative Selection algorithm has been applied to intrusion detection problems in the past and has received some criticism regarding its "scaling problems" [54]. Gonzalez and Dasgupta argue that their new representation is the key to avoiding the scaling issues of the original algorithm. Their results show that for their particular problem the SOM algorithm and their own algorithm are comparable overall. Another comparison of SOM to a novel AIS based approach is performed by Greensmith *et al.* [35]. Their comparison is of Kohonen's original SOM versus an algorithm based on a cell of the human immune system called the dendritic cell. Their results have shown that the Dendritic Call algorithm performed statistically significantly better than SOM in a port scanning scenario.

Lei and Ghorbani [66] compare SOM to an *improved competitive learning* network (ICLN) which is based on a single-layer neural network. The authors state that the ICLN approach is comparable to results obtained by a SOM, however at a dramatically smaller computational overhead.

Wang *et al.* [104] compare Kohonen's original SOM algorithm with HMM. Their findings are that HMM is better than SOM for one type of dataset (Sendmail), while for another (Live FTP) both approaches have comparable results. Nevertheless the HMM approach requires a considerable amount of time in comparison to the SOM approach, making the SOM more suitable for real-world applications.

Amini *et al.* [2] compare SOM with two types of ART algorithms. The results of their work show that their ART algorithms perform better, both in terms of speed as well as detection accuracy. Durgin and Zhang [20] also perform comparison of SOM and ART methods for intrusion detection. Their version of the ART algorithm incorporates fuzzy logic and is said to be significantly more sensitive than the tested SOM approach.

Sarasamma and Zhu [93] compare their hyperellipsoidal SOM against a number of other intrusion detection approaches, including ART, RBF, MLP, ESOM and many others. They conclude that by using the combination of their own version of the SOM algorithm with the ESOM method gives excellent results in comparison to the other tested techniques.

### 3.1.2.  Intrusion and Anomaly Alerts

Intrusion detection systems suffer from a number of disadvantages. One of the major issue with such systems is the amount of alerts that such systems generate. In order for an IDS to provide a manageable amount of alerts that can be reasonably dealt with by an administrator, a number of alert filtering techniques have been developed. Some of those incorporate the SOM algorithm for various purposes.

Faour *et al.* [21] employ SOM and Bayesian belief networks in order to automatically filter intrusion detection alarms. SOM in this case is used to cluster attack and normal scenarios, with the Bayesian method used as a classifier. Their system is able to filter 76% of false positive alarms. Faour *et al.* [22] introduce the combination of SOM and *growing hierarchical* SOM (GHSOM) for the purpose of interesting pattern discovery in terms of possible real attack scenarios. They find that the GHSOM addresses two main limitations of SOM, namely static architecture and lack of hierarchical representation of relations of the underlying data. Shehab *et al.* [96] extend the previous model by introducing a decision support layer to enable administrators to analyse and sort out alarms generated by the system. They have also shown empirically that GHSOM has the potential to perform better than the rigid-structured original SOM.

Another drawback of existing IDSs is the lack of meaning of generated alerts. Any logical connection between generated alarms is usually omitted. For this purpose a number of researchers started looking into intrusion alert correlation. SOM has also been used within this area, most notably by Smith *et al.* [98] and Xiao and Han [106]. Smith and colleagues [98] develop a two stage alert correlation model where in the first stage individual attack steps are grouped together and in the second stage a whole attack is grouped together from the groups generated within the first stage. In their work SOM is used for the first stage. Experiments however deem the SOM noticeably worse than an algorithm proposed by the authors. Xiao and Han [106] on the other hand create a system which correlates intrusion alerts into attack scenarios. The authors use an improved ESOM, which enables evolution of the network and fast incremental learning. The output of the system are visual attack scenarios presented to an administrator.

### 3.1.3.  Visualisation

Due to the SOM algorithm's capability of visualising multidimensional data in a meaningful way, its use lends itself ideally to its application in visualising computer security problems. Gonzalez *et al.* [34] use this ability to visualise the self/non-self space that they use for anomaly detection. This visualisation presents a clear discrimination of the different behaviours of the monitored system. Hoglund *et al.* [41, 40] on the other hand employ visualisation of user behaviour. In their work various host based signals are used for monitoring of users. A visual representation is subsequently presented to administrators in order for them to be able to make an informed decision in case of unacceptable user behaviour.

Kumar and Devaraj [61] use SOM along with BPN for visualisation and classification of intrusions. In this system the SOM helps to visualise and study the characteristics of each input feature. Jirapummin *et al.* [46] also use SOM however in this case for visualisation of malicious network activities using a U-Matrix. In their system this enables to visually distinguish between different types of scanning attacks. Xiao and Han [106] use SOM as a correlation technique that produces visualisations of whole attack scenarios.

Girardin and Brodbeck [30] and later Girardin [29] develop a system that takes away the burden of an administrator to look through logs of audit data. The SOM algorithm is employed to classify events within such logs and present these events in a meaningful way to an administrator. The authors have successfully developed tools to monitor, explore and analyse sources of real-time event logs using the SOM algorithm. In [29], the author uses the developed tools in order to monitor a dataset with known attacks. The paper concludes by stating that the tools are an effective technique for the discovery of unexpected or hidden network activities. Nevertheless the author also states that after analysing network traffic at the protocol level, it is apparent that such information might not be encompassing enough to make complex patterns apparent. A more complex and varied data would possibly enable this.

Yoo and Ultes-Nitsche [107] use SOM for visualisation of computer viruses within Windows executable files. Yoo has found that patterns representing virus code can be found in infected files using the SOM visualisation technique (U-Matrix). Their technique discovered a DNA-like pattern across multiple virus variations.

### 3.1.4.  Binary Code Analysis

As mentioned in previous section, SOM algorithm has also been used for the analysis of binary code. Yoo and Ultes-Nitsche [107] analysed windows executables by creating maps of EXE files before and after an infection by a virus. Such maps have been subsequently analysed visually and found to have contained patterns, which can be thought of as virus masks. The author states that such masks can be used in the future for virus detection in a similar manner to current anti-virus techniques. The difference being that a single mask could detect viruses from a whole virus family rather than being able to find only a single variant. In 2006, Yoo and Ultes-Nitsche [109] extend their work by testing their proposed SOM based virus detection technique on 790 virus-infected files, which includes polymorphic as well as encrypted viruses. Using their approach the system is able to detect 84% of all infected files however at a quite high false positive rate of 30%. The authors conclude that this technique complements existing signature based anti-virus systems by detecting unknown viruses. Yoo and Ultes-Nitsche [108] also look at packet payload inspection using their binary code analysis technique. In this case the system is implemented as part of a firewall.

Payer *et al.* [84] investigate different statistical methods, including the SOM, for the purpose of polymorphic code detection. They have observed three different techniques, looking only at packet payload without any other additional information. Their conclusion is that SOM does not provide detection rates on par with their other neural network technique. Bolzoni *et al.* [9] also look at payload monitoring using SOM by employing a two-tier architecture intrusion detection system. They state that the SOM enables dramatic reduction of profiles, necessary for detection, to be created using this system.

Buffer overflow attack detection has also been tackled, namely by Rhodes *et al.* [90] and Ramadas *et al.* [89]. Rhodes and colleagues [90] monitor packet payloads using a multilayer SOM in order to detect buffer overflows against a DNS server. Ramadas and colleagues [89] perform detection using SOM as part of an existing real-time system. Their system is successful at detecting buffer overflow attack for the Sendmail application. Their conclusion is that the SOM algorithm is particularly suitable for buffer overflow detection.

### 3.1.5.   Attacks and Vulnerabilities

Due to SOM's capabilities also as a classification algorithm, a number of researchers have shown its use for the purpose of attack and vulnerability classification. This vital aspect of intrusion detection enables administrators quickly asses the importance of an alert and thus be able to make an informed decision about what action to take.

DeLooze [16] uses the SOM algorithm in order to classify the database of common vulnerabilities and exposures (CVE), based on their textual description. The author argues that attacks that are in the general neighbourhood of one another can be mitigated by similar means. Their system is able to create a map of the common attack classes based on the CVE database.

Venter *et al.* [101] attempt to tackle the same problem as DeLooze. They also employ the SOM algorithm for the purpose of clustering the CVE database. They state that the advantage of having such a system is to be able to assess vulnerability scanners. Their system distinguishes 7 attack classes, rather than 4, as is the case in DeLooze's work. Their findings show that there is lack of standardisation of naming and categorisation of vulnerabilities, making it difficult to assess and compare vulnerability scanners.

Pan and Li [83] use SOM in combination with RBF in order to classify novel attacks. Their system is largely an IDS which directly classifies an anomaly into one of a number of predefined attack categories.

Doumas *et al.* [19] attempt to recognise and classify viruses using a SOM and a BPN. The authors have analysed DOS based viruses. They find that the BPN requires fewer steps than the SOM in order to obtain acceptable results, on the other hand the SOM does not require any class information and is still able to obtain clusters of similar patterns.

DeLooze [17] employ an ensemble of SOM networks for the purpose of an IDS as well as for attack characterisation. Genetic algorithms are used for attack type generation, subsequently employed as part of an IDS that is able to discriminate the type of attack that has occurred.

### 3.1.6.   Email and Spam

An important aspect of software security that is increasingly putting burden on businesses and individuals is the issue of spam and malicious email messages. Some authors have approached to tackle the issue of malicious code detection in email attachments, such as the work of Yoo and Ultes-Nitsche [108]. They look at packet payload inspection using their binary code analysis technique for SMTP traffic. Their system is said to be able to detect a variety of existing as well as novel worms and viruses, however policies and probabilities used to tune the system still need significant development.

Others attempt to solve the issue of spam emails with the help of the SOM algorithm. For example the work of Ichimura *et al.* [44] attempts to classify spam emails based on the results of an open source tool called SpamAssasin. Their system categorises spam into different groups, from which rules are subsequently extracted in order to aid SpamAssasin with detection. This rule extraction is performed using agents and genetic programming. Their system is able to improve the detection of spam emails, however with some false positives.

Cao *et al.* [11] also attempt to solve the problem of spam emails. They use a combination of PCA and SOM to perform this task. PCA is used in order to select the most relevant features of emails to be fed to a SOM. The SOM is used to classify the observed email into two categories, spam or normal. Their results show a performance of almost 90% in filtering email.

Luo and Zincir-Heywood [71] introduce a SOM based sequence analysis for spam filtering. Their system also uses a *k-Nearest Neighbour* algorithm as a classifier. A comparison of their system with a *Naive Bayesian* filter is performed and the SOM method is found to achieve better results. The authors however state that the efficiency of the SOM approach is not completely elaborated.

As mentioned earlier, Ramadas *et al.* [89] develop a module for an intrusion detection system which besides other protocols, is able to monitor SMTP traffic. Their system is able to successfully detect buffer overflow attacks.

### 3.1.7.   Other Software Security Problems

Two more pieces of research work are worthy of mentioning in this section. First of all the work of Chan *et al.* [13], who propose a web policing proxy able to dynamically block and filter Internet contents. Their system employs Kohonen's algorithm for performing realtime textual classification with a classification rate of 64%. Their work is the first instance of using the SOM algorithm for web application security.

The other research work deals with access control. Weipel *et al.* [105] introduce a SOM based access control technique to determine access rights to documents based on their content. The system is also able to classify the document's access levels and whether incorrect settings are assigned to documents due to SOM's clustering and classification capabilities.

### 3.2.   Hardware Security

In this section, we will focus on the use of the SOM algorithm in the more tactile areas of of security. Kohonen's algorithm hasn't seen as much attention in this area as in software security, nevertheless some areas, such as biometrics, strongly benefit from the algorithm's clustering and classification properties.

### 3.2.1.  Biometrics

In biometrics various feature recognition techniques are necessary in order to classify visual, auditory and haptic signals for the purpose of security and authentication. Due to SOM's success in the image and vision recognition areas, the algorithm has been applied to a number of biometric systems. For example Herrero-Jaraba *et al.* [39] use the SOM algorithm for human posture recognition in video sequences for the purpose of physical and personal security. Kumar *et al.* [60] on the other hand use SOM for face recognition. They use the SOM algorithm along with PCA. Monteiro *et al.* [79] also use SOM for facial recognition, nevertheless, in this case, independent of facial expressions. The authors compare their SOM based approach to other neural based approaches such as MLP and RBF and have shown that they have obtained comparable results. Khosravia and Safabakhsha [53] use a time adaptive SOM for human eye-sclera detection and tracking. Their experiments show that their system could be used for real-time detection. Bernard *et al.* [7] use SOM for fingerprint pattern classification. The authors state that this method provides an efficient way of classifying fingerprints. Their system provides 88% classification on a standard dataset, which is a good result, nevertheless one which should be increased to at least 98% in order to be comparable to other best approaches. Shalash and Abou-Chadi [95] also use SOM for fingerprint classification. Their system uses a multilayer SOM, which achieves 91% detection accuracy on the same dataset as used in Bernard's work. Martinez *et al.* [74] look at biometric hand recognition using a supervised and unsupervised SOM with LVQ. Their system performs well in comparison to other methods due to low false positives. The authors state that based on these results, biometric hand recognition can be used for low to medium-level security applications.

### 3.2.2.  Wireless Security

The field of wireless networking and its security is currently a hot topic in computer science. Decreasing costs of wireless technologies enable widespread use of mobile networks in all aspects of our lives. Some work using the SOM algorithm has also been performed in various branches of wireless networking.

The work of Boukerche and Notare [10] for example looks at fraud in analogue mobile telecommunication networks. Their system is able to identify a number of malicious users of mobile phones based on a number of telecommunication indicators such as network characteristics and temporal usage. The authors state that the performance of their detector is able to reduce profit loss of phone operators to between 1% and 10% depending on the performance of their neural model.

Grosser *et al.* [36] also look at fraud in mobile telephony. The authors observe unusual changes in consumption of mobile phone usage. In their system SOM is used for pattern generation of various types of calls. These patterns are then used to build up a profile of a user, later used as a baseline for unusual behaviour detection.

Kumpulainen and Hatonen [62] develop an anomaly based detection system that looks at local rather than global thresholds, which depend on local variation of data. Their experiments are performed on server log and radio interface data from mobile networks. The authors state that their local method provides interesting results compared to a global method. Mitrokotsa *et al.* [78] introduce both an intrusion detection and prevention system. Emergent SOM is used for both visualisation and intrusion detection and a watermarking technique is used for prevention. Their system is implemented in every node of a mobile ad-hoc (MANET) network in such a way that each node communicates between each other in order to compose an IDS for the network. Using ESOM a feature map is created for each node as well as the whole network. In their system the visualisation of the ESOM is exploited for the purpose of intrusion detection.

Avram *et al.* [3] use SOM for attack detection in wireless ad-hoc networks. Their system monitors network traffic on individual nodes of the network and anlyses it using the SOM algorithm. A number of routing protocols for MANET networks are monitored and it is shown that high detection rates can be achieved to detect different types of network attacks with low amount of false positive alerts.

It is interesting to note that to our knowledge, the SOM algorithm has not been used thus far for security purposes in other areas of wireless communications, notably within the Bluetooth and Radio Frequency Identification (RFID) areas. This is surprising as with the increase of activity in both of those fields, especially RFID, the need for intrusion detection and RFID chip monitoring systems is apparent.

### 3.2.3.  Smartcards

An interesting application of the SOM algorithm can be found in [88]. Quisquater [88] uses the SOM with traditional correlation techniques in order to monitor execution instructions of a smart card processor. The author develops an attack that is able to eavesdrop on processed data by monitoring the electric field emitted by the processor. The author concludes that this type of attack will become increasingly more relevant in the future and should be investigated further.

### 3.3.    Other Security Areas

Numerous other areas of computer security exist. In this section we have selected a subset of those, where the SOM algorithm has been used for a substantial amount of work performed by the developed research work.

### 3.3.1.  Cryptography

Jamzad and Kermani [45] propose that different images have different abilities to hide a secret message within them. They propose a method for finding steganographically suitable images using a combination of a Gabor filter and the SOM. In their system the SOM is used to determine the most suitable image, based on the data supplied to it by the Gabor filter. In contrast Oliveira *et al.* [82] use SOM as a clustering and categorisation tool for attacking cryptosystems.

### 3.3.2.   Forensics

Forensics can be thought of as a data mining issue. From this point of view a SOM is an ideal candidate for understanding or extracting unknown information form various data sources.

Beebe and Clark [5] state that an issue in forensics text string searching is the retrieval of results relevant to digital investigation. The authors propose the use of SOM for the purpose of post-retrieval clustering of digital forensic text. Experimental results show favourable results for their method, nevertheless a number of issues pertain. Firstly the issue of scale and secondly whether such clustering does indeed help investigators.

Fei *at al.* [23, 24] also use SOM as a decision support tool for computer forensic investigations. In this case SOM is used for more efficient data analysis, utilising the algorithm's visualisation capabilities. Anomalous behaviour of users is visualised and better understanding of underlying complex data is enabled in order to give investigators better view of the problem at hand.

Oatley *et al.* [80] provides a thorough analysis and discussion of existing techniques used for forensic investigation of crimes by police. The authors describe the use of Kohonen's SOM across a variety of both digital and non-digital forensics in order to help investigators solve crimes.

### 3.3.3.   Fraud

Kohonen's SOM has been used for fraud detection on a number of occasions. As already mentioned previously the work of Boukerche and Notare [10] looks at fraud in analogue mobile telecommunication networks. Their system is able to identify a number of malicious users of mobile phones based on a number of telecommunication indicators such as network characteristics and temporal usage.

Grosser *et al.* [36] also look at fraud in mobile telephony. The authors observe unusual changes in consumption of mobile phone users. In their system SOM is used for pattern generation of various types of calls. These patterns are then used to build up a profile of a user, later used as a baseline for unusual behaviour detection.

Quah and Sriganesh [86, 87] use SOM for real-time credit card fraud detection. Their SOM based approach allows for better understanding of spending patterns by deciphering, filtering and analysing customer behaviour. The SOM's clustering abilities allow the identification of hidden patterns in data which otherwise would be difficult to detect.

### 3.3.4.   Home Security

Oh *et al.* [81] propose the use of the SOM algorithm as part of a home gateway to detect intrusions in real-time. At the moment their system is a traditional SOM based IDS in nature, nevertheless their uniqueness is in an architecture which takes into account various home based appliances interconnected by a gateway and monitored by the proposed IDS.

### 3.3.5. Privacy

Han and Ng [37] extend the SOM algorithm in such a way that when used for various machine learning and data mining purposes, the algorithm preserves the privacy of parties involved. The authors propose protocols to address privacy issues related to SOM. In their work they prove that such protocols are indeed correct and privacy conscious.

## 4. Discussion

From the overview of literature of SOM based security research we can draw a number of conclusions. The SOM algorithm is a successful artificial intelligence technique that is applicable across a wide variety of security problems. The algorithm's strengths lie mainly in clustering and visualisation of complex, highly dimensional data that are otherwise difficult to understand. SOM's clustering capabilities enable it to be used as an effective anomaly detector which can be used in real-time systems, depending on the problem at hand. On its own, the algorithm does achieve good performance in many problem areas, however other algorithms, especially ones which are suited for classification, perform better. For this reason the SOM algorithm performs best when coupled with other approaches such as SVM, HMM or PCA or when extended to tackle a particular problem. Selection of ideal parameters for generation of SOM features maps is still a problematic area, nevertheless this issue is tackled by some extended SOM methods.

Looking at areas of security in which the algorithm has been applied in the past, it is apparent that anomaly detection dominates the field. Many other software security problems have been tackled with the help of the SOM as well, nevertheless numerous areas of security have not yet been approached from a SOM point of view. For example the issue of bots and botnet detection, malware classification or radio frequency identification, could benefit from the clustering and visualisation capabilities of the algorithm. Issues such as insider threat and copyright are also thus far to be looked at. Due to SOM's general machine learning nature and numerous advantages, its application in all of the above mentioned security areas could undoubtedly benefit the security areas' research portfolios.

The issue of SOM performance deserves a discussion on its own. Kohonen originally based his SOM algorithm on the biological property of somatotopic map creation in the human brain as described in section 2.1.. It is a known fact that a mammalian brain is a highly parallel structure that is able to process vast amounts of data at the same time. The fact that the SOM algorithm comprises of, usually, a 2D layer of nodes, each of which performing a computation at every step of the algorithm's operation, the usefulness of machines able to perform parallel computation is undisputed. In the last few years, the field of general purpose processors has slowly started to shift towards these types of computational architectures. The introduction of multi-core general purpose CPU's and inclusion of more specialised highly multi-core architectures, such as the CELL/B.E., into home entertainment devices, marks a step forward for algorithms that benefit from parallelism. The SOM algorithm is one of such algorithms and with the increase of parallelism, issues of computational overhead and thus limitations due to complexity of desired map will increasingly be eliminated. This, coupled with the general success of the algorithm within the security field, evidence of sustained interest in extending the work proposed by Kohonen and areas of security still untouched by the algorithm, suggest that still many possibilities lie ahead for researchers in applying SOM and its incarnations to various security problems.

## 5.  Conclusion

In this chapter we have introduced a biologically inspired algorithm called the SelfOrganising Map. This algorithm has been used in over a hundred security related research works and has achieved a substantial interest due to its strengths and capabilities as a tool for visualisation, clustering and classification. The area of software security and in particular intrusion detection has seen the largest amount of interest from within research work conducted with the SOM algorithm. Some experimental evidence has shown that the algorithm performs on par with other established computational intelligence techniques in terms of detection and computational overhead performance. Our review of literature has also revealed that some unique uses of the algorithm opened up areas of security which have not been tackled in a similar way before, such as anomaly based detection and classification of viruses.

Some areas of security have as of now been untouched by the algorithm even though the algorithm's capabilities lend themselves ideally for such use. Examples of such areas are radio frequency identification and bot detection.

The original Kohonen's algorithm has been developed over two decades ago. Since then numerous incarnations, versions and adjustments have been proposed, to exploit or improve the functionalities of the algorithm, with encouraging results. The combination of the algorithm with other machine learning approaches have also shown great results. With the increasingly multi-threaded nature of computing in terms of multi-core computing architectures, such as the CELL/B.E. processor, the authors feel that the SOM algorithm and its various incarnations have a bright future.